%% file: main.tex
\documentclass[conference]{IEEEtran}
\IEEEoverridecommandlockouts
\usepackage{cite}
\usepackage[T1]{fontenc}
\usepackage{amsmath,amssymb,amsfonts}
\usepackage{booktabs}
\usepackage{algorithmic}
\usepackage{graphicx}
\usepackage{textcomp}
\usepackage{xcolor}
\usepackage{array}
\usepackage{multirow}
\usepackage{url}
\def\BibTeX{{\rm B\kern-.05em{\sc i\kern-.025em b}\kern-.08em
    T\kern-.1667em\lower.7ex\hbox{E}\kern-.125emX}}

\newcommand{\best}[1]{\textbf{#1}}
\newcommand{\second}[1]{\textit{#1}}

\include{preamble}

\begin{document}

\AddToHookNext{shipout/foreground}{%
  \begin{tikzpicture}[remember picture,overlay]
    \node[
      anchor=north,
      font=\normalsize
    ] at ([yshift=-0.8cm]current page.north)
    {Accepted as a conference paper at IEEE BigData 2026};
  \end{tikzpicture}%
}
\title{DISC: Domain-Integrated Similarity Cost for Time-Series Dataset\\}

\author{
\IEEEauthorblockN{Yen-Ku Liu}
\IEEEauthorblockA{
\textit{National Yang Ming Chiao Tung University}\\
yenkuliu.cs15@nycu.edu.tw
}
\and
\IEEEauthorblockN{Hongjie Chen}
\IEEEauthorblockA{
\textit{Dolby Laboratories}\\
hongjie.chen@dolby.com
}
\and
\IEEEauthorblockN{Franck Dernoncourt}
\IEEEauthorblockA{
\textit{Adobe Research}\\
dernonco@adobe.com
}
\and
\IEEEauthorblockN{Ryan Rossi}
\IEEEauthorblockA{
\textit{Adobe Research}\\
ryrossi@adobe.com
}
}

\maketitle

\begin{abstract}
The growth of time-series datasets and foundation models has made dataset comparison a central part of time-series workflows.
Assessing source-target transfer before model adaptation requires similarity estimates that can summarize transfer-related dataset relationships without exhaustive source-target evaluation over every candidate pair.
Many metrics have been proposed for measuring similarity between time-series datasets, but they capture different dataset relationships and vary in how well they align with downstream utility across tasks.
We evaluate this gap by measuring the correlation between source-target similarity scores and downstream forecasting errors in out-of-distribution and transfer-learning settings.
We propose Domain-Integrated Similarity Cost (DISC), a learnable aggregation framework that uses calibration transfer observations to select, weight, and combine similarity metrics for different downstream tasks.
By integrating complementary metric values, DISC better distinguishes source-target dataset differences that matter for downstream behavior.
After calibration, DISC scores source-target pairs using only their similarity-metric values and yields higher alignment than single-metric references on the reported split.
These results show that time-series dataset similarity is task-dependent, and that learned metric integration provides a stronger proxy for downstream transfer behavior than manually selecting one similarity metric.
\end{abstract}

\begin{IEEEkeywords}
time-series dataset similarity, transfer learning, foundation models, correlation evaluation
\end{IEEEkeywords}

\section{Introduction}
Time-series data appear in energy, transportation, weather, healthcare, finance, and web-traffic systems~\cite{zhang2017stock,che2018recurrent,nguyen2023climax,lam2023learning,wu2022timesnet}.
As time-series research has expanded, dedicated resources have been developed for classification~\cite{bagnall2018uea}, anomaly detection~\cite{paparrizos2022tsb}, forecasting~\cite{hahn2023time,wang2026deep,aksu2024gift}, and foundation-model development~\cite{ansari2024chronos,das2023decoder,woo2024unified}.
Recent foundation models for time-series forecasting make it possible to adapt pretrained models across heterogeneous domains~\cite{ansari2024chronos,kottapalli2025foundation,ma2024survey}.
This growth makes relevant source-target transfer assessment important for downstream time-series tasks, particularly when models rely on transfer, adaptation, or external data reuse~\cite{han2025retrieval,ning2026ts,jin2024time,liu2024unitime,weber2021transfer}.
In transfer-learning and adaptation settings, source-target compatibility can affect whether fine-tuning provides useful signal~\cite{pan2009survey,ehrig2024impact,weber2021transfer}, while a mismatched pair can degrade target-domain performance or introduce negative transfer~\cite{pan2009survey,wang2019characterizing,zhang2022survey}.

\begin{figure}[t]
\centering
\includegraphics[width=\linewidth]{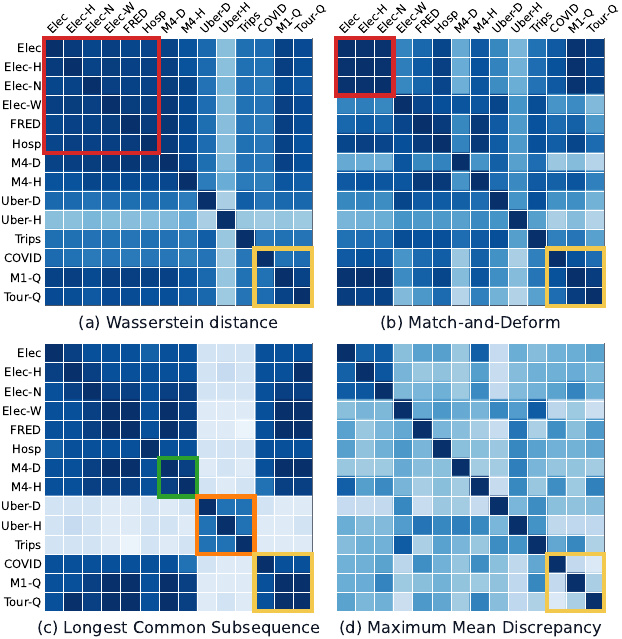}
\caption{Pairwise dataset distances under four similarity metrics. Darker cells indicate smaller distance and therefore higher metric-implied similarity. The panels illustrate why single-metric similarity is insufficient: WSD captures a broad related group, MAD sharpens only part of that relation, DBA+LCSS reveals additional temporal-shape similarities, and MMD rejects a spurious group that the other metrics rank as close. This motivates learning an integrated similarity cost that combines complementary evidence while avoiding over-reliance on any single metric.}
\label{fig:intro_motivation}
\end{figure}

As a result, quantifying similarity between time-series datasets becomes crucial.
However, this utility depends on what relationship a metric captures: two datasets may be close in marginal distribution, temporal shape, representative pattern, or alignment structure, and each view can be more or less informative for a particular downstream task\cite{paparrizos2024survey}.
Figure~\ref{fig:intro_motivation} illustrates how different single similarity metrics expose different parts of the dataset relationship.
In Fig.~\ref{fig:intro_motivation}(a)--(b), both Wasserstein distance (WSD) and Match-and-Deform (MAD) identify the electricity-related datasets as similar, as indicated by the red outlines.
MAD further sharpens this relation by revealing that \texttt{electricity}, \texttt{electricity\_hourly}, and \texttt{electricity\_nips} form a tighter core within the broader electricity group.
However, MAD does not clearly mark \texttt{m4\_daily} and \texttt{m4\_hourly} as similar, whereas Dynamic Time Warping Barycenter Averaging with Longest Common Subsequence (DBA+LCSS) gives this green-outlined pair a clearer low-distance relation in Fig.~\ref{fig:intro_motivation}(c).
DBA+LCSS also gives the clearest similarity result for the orange-outlined mobility datasets \texttt{uber\_tlc\_daily}, \texttt{uber\_tlc\_hourly}, and \texttt{vehicle\_trips\_wm}, while WSD and MAD leave the pairs involving \texttt{uber\_tlc\_hourly} comparatively distant.
At the same time, not every low-distance value should be trusted uniformly.
For the yellow-outlined semantically mismatched group \texttt{covid\_deaths}, \texttt{m1\_quarterly}, and \texttt{tourism\_quarterly}, WSD, MAD, and DBA+LCSS all rank the three pairwise relations as relatively close, but Maximum Mean Discrepancy (MMD) ranks them as distant.
The yellow-outlined case shows that a low distance under one metric can be misleading when other metrics capture a conflicting dataset relationship.
Together, these examples motivate a learnable framework that integrates metric evidence according to downstream utility, rather than relying on any single metric that happens to mark a pair as similar.

These metric-level differences matter because dataset similarity is used beyond pairwise comparison: it can support transfer assessment~\cite{sun2025quantifying,zhang2026unified,yao2025estimating}, dataset visualization and inspection~\cite{sun2025quantifying,paparrizos2024survey}, benchmark curation~\cite{hahn2023time,aksu2024gift}, and analysis of cross-domain generalization~\cite{ehrig2024impact}.
Since evaluating every source-target transfer setting is costly, dataset similarity has been used to prioritize source-target evaluations before downstream adaptation~\cite{sun2025quantifying,zhang2023similarity,ding2024model}.
A useful similarity score should therefore vary with downstream transfer behavior, especially because source-target similarity can affect forecasting performance across zero-shot and fine-tuned settings~\cite{ehrig2024impact,jain2023data}.
We therefore treat time-series dataset similarity as a correlation-alignment problem between source-target similarity scores and downstream task errors, which also explains why no single metric should be expected to dominate across all tasks.
These observations suggest that dataset similarity should not be treated as a fixed metric choice.

To address the gap between single-metric similarity and downstream task behavior, this paper proposes Domain-Integrated Similarity Cost (DISC), a learnable aggregation framework for time-series dataset similarity.
DISC uses observed transfer results on calibration datasets to learn metric mixtures that can then be applied to other source-target pairs using only dataset similarity values.
The base mixture forms the primary similarity score from standardized metric values.
The prior mixture forms a rank-based value from source-wise metric ranks.
A learned sigmoid gate uses the prior value to modulate the base score.
This design lets DISC emphasize metrics that provide useful cost value while using rank-based information as supplementary source-wise evidence.
The resulting workflow integrates complementary metric evidence into a similarity score intended to better separate datasets whose downstream behaviors differ.

We evaluate DISC across out-of-distribution and transfer-learning forecasting settings with multiple similarity configurations.
On the reported split, DISC increases the average test correlation by about 10\% across the four task settings.

Our contributions are:
\begin{itemize}
    \item DISC integrates complementary dataset similarity metrics, which enables the learned similarity score to better discern source-target differences associated with downstream transfer behavior.
    \item DISC is flexible and task-adaptable to different downstream settings, using learnable aggregation parameters to adjust metric composition and weighting toward improved downstream-aligned similarity estimation.
    \item The learned metric weights provide an interpretable view of which dataset relationships are useful for a given model and task setting.
\end{itemize}

\section{Related Work}
\noindent\textbf{\textit{Time-series similarity metrics.}}
Existing similarity metrics capture different views of time-series dataset relationships.
Dataset-level distances compare empirical collections directly.
\textit{Maximum Mean Discrepancy (MMD)} measures distribution discrepancy using kernel mean embeddings~\cite{gretton2012kernel}, \textit{Optimal Transport (OT)} measures minimum transport cost between empirical samples~\cite{cuturi2013sinkhorn,gabriel2019computational,alvarez2020geometric}, \textit{Wasserstein distance} models each dataset through a Gaussian approximation and relates the resulting distance to downstream behavior~\cite{chen2025measuring}, and \textit{Match-and-Deform (MAD)} jointly optimizes sample matching and temporal deformation for time-series domain adaptation~\cite{painblanc2023match}.
Series-level distances compare individual sequences and can be lifted to dataset comparison by first reducing each dataset to a representative sequence.
\textit{Dynamic Time Warping (DTW)} captures temporal warping~\cite{sakoe1978dynamic,cuturi2017soft}, \textit{Longest Common Subsequence (LCSS)} captures subsequence matches while tolerating unmatched regions~\cite{hirschberg1977algorithms,vlachos2002discovering}, and \textit{Euclidean Distance (ED)} compares aligned shapes~\cite{faloutsos1994fast,ding2008querying}.
Reducers such as \textit{Dynamic Time Warping Barycenter Averaging (DBA)}~\cite{petitjean2011global} and \textit{Principal Component Analysis (PCA)}~\cite{pearson1901liii} make these sequence distances usable as dataset-level configurations, although they may discard within-dataset heterogeneity.
Because these metric families emphasize different properties, their suitability can vary across datasets and downstream tasks~\cite{d2024beyond}.

\noindent\textbf{\textit{Time-series foundation models and transfer.}}
Time-series foundation models reuse broad temporal representations across domains and tasks, including Lag-Llama for probabilistic lagged-sequence forecasting~\cite{rasul2023lag} and Time-MoE for scaled mixture-of-experts forecasting~\cite{shi2025time}.
Other foundation models, such as Chronos~\cite{ansari2024chronos}, TimesFM~\cite{das2023decoder}, Moirai~\cite{woo2024unified}, MOMENT~\cite{goswami2024moment}, and LLM-reprogramming approaches, further show that sequence modeling can generalize across diverse time-series collections~\cite{jin2024time,liu2024unitime}.
These models make dataset reuse important because source-target compatibility can provide adaptation value or harmful bias, and retrieval-augmented and transferability-aware forecasting methods further show that performance depends on how external datasets are reused~\cite{han2025retrieval,ning2026ts,yao2025estimating,zhang2026unified}.
This motivates evaluating similarity metrics by how well they align with downstream forecasting behavior, rather than only by mathematical or visual plausibility.

\begin{figure*}[t]
\centering
\includegraphics[width=0.95\textwidth]{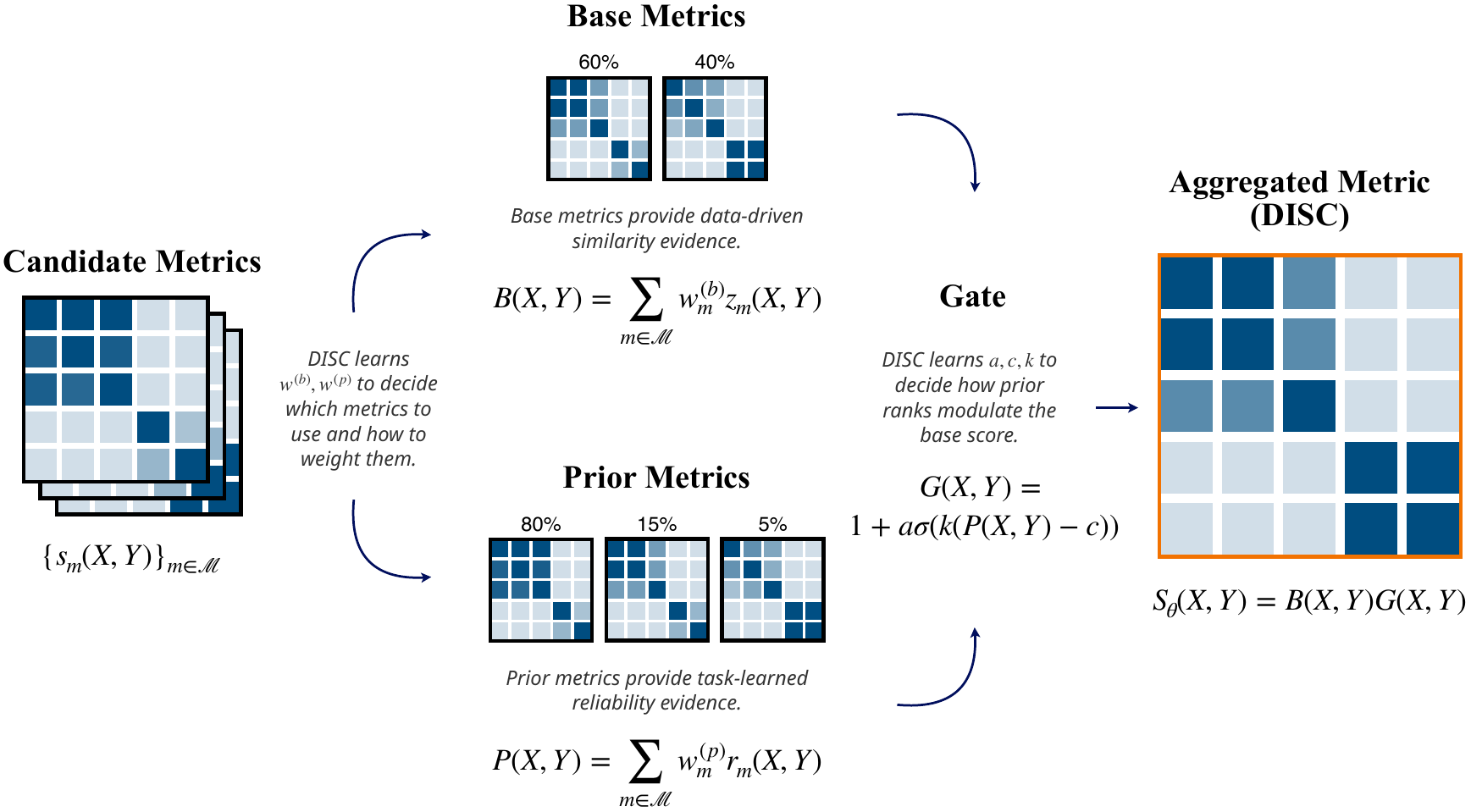}
\caption{Overview of Domain-Integrated Similarity Cost (DISC). Given candidate similarity metrics $\{s_m(X,Y)\}_{m\in\mathcal{M}}$, DISC learns a task-specific base mixture $B(X,Y)$ from standardized metric values $z_m(X,Y)$ and a prior mixture $P(X,Y)$ from source-wise metric ranks $r_m(X,Y)$. A learned gate $G(X,Y)$ uses the prior mixture to modulate the base mixture and produce the final score $S_\theta(X,Y)=B(X,Y)G(X,Y)$. The base and prior mixtures are detailed in Section~\ref{sec:method_selection}, and the gated integration is detailed in Section~\ref{sec:method_gate}.}
\label{fig:disc_architecture}
\end{figure*}

\section{Preliminaries}
\label{sec:preliminaries}
Let $X=\{x_i\}_{i=1}^{N_X}$ and $Y=\{y_j\}_{j=1}^{N_Y}$ denote two time-series datasets.
Each sequence $x_i=(x_{i,1},\ldots,x_{i,T_i})$ and $y_j=(y_{j,1},\ldots,y_{j,T_j})$ may have a different length, with observations in $\mathbb{R}^d$.
Here, $N_X$ and $N_Y$ are the numbers of sequences in the two datasets, $T_i$ and $T_j$ are sequence lengths, and $d$ is the observation dimension.
Time-series dataset similarity aims to define a dissimilarity function $s(X,Y)$, where smaller values indicate higher similarity.

We consider two classes of similarity methods.
Dataset-level metrics compare two datasets directly as $s_m(X,Y)=m(X,Y)$, where $m(\cdot, \cdot)$ denotes a metric whose inputs are time-series datasets.
However, not all useful time-series distances are defined between datasets.
Some distances compare only individual time series, while our target object is the similarity between two time-series datasets.
We therefore use a reducer to summarize each dataset into a representative sequence, making series-level distances applicable to dataset-level comparison.
Reducer-based series-level metrics first map each dataset to a representative sequence using a reducer $R$, then apply a series-level distance $d$ as $s_{R,d}(X,Y)=d(R(X),R(Y))$.
Here, $s_{R,d}$ denotes the dataset-level distance induced by reducer $R$ and series-level distance $d$.
This abstraction covers the dataset-level and reducer-based similarity configurations used in our experiments.

Let $\mathcal{U}=\{X_1,\ldots,X_n\}$ denote the dataset universe.
Here, $n$ is the number of datasets and $\mathcal{M}$ is the set of candidate similarity methods.
For each similarity method $m\in\mathcal{M}$, the pairwise scores form a distance matrix $\mathbf{S}^{(m)}\in\mathbb{R}^{n\times n}$, where $\mathbf{S}^{(m)}_{ab}=s_m(X_a,X_b)$.
The downstream evaluation produces directed error entries, represented as $\mathbf{E}_{ab}=E(X_a,X_b)$ when the entry is available, where $E(X_a,X_b)$ is the error obtained by using $X_a$ as the source dataset and evaluating on $X_b$ as the target dataset.

\section{DISC: Domain-Integrated Similarity Cost}
\label{sec:method}
In this section, we introduce the core ideas of Domain-Integrated Similarity Cost (DISC).
DISC is designed to learn how multiple time-series dataset similarity metrics should be selected, weighted, and combined for a given downstream task.
Instead of assuming that one metric is always reliable, DISC treats each metric as a candidate signal that may capture a different aspect of the source-target dataset relationship.
Figure~\ref{fig:disc_architecture} provides an overview of these components and how they produce the final dataset-similarity score.
We first define the source-target problem setting in Section~\ref{sec:method_problem}, describe base and prior metric selection in Section~\ref{sec:method_selection}, introduce the gated integration mechanism in Section~\ref{sec:method_gate}, and present the training objective and procedure in Section~\ref{sec:method_training}.

\subsection{Problem Setting}
\label{sec:method_problem}
In this subsection, we define the source-target setting and the overall form of the DISC score.
The input is a collection of source-target distance matrices and downstream error entries over the same dataset universe.
The output is a scalar score for each ordered pair $(X,Y)$.
At a high level, DISC represents the final similarity score as
\begin{equation}
S_\theta(X,Y)=B(X,Y)G(X,Y),
\label{eq:disc_overview}
\end{equation}
where $S_\theta(X,Y)$ denotes the final learned DISC score for the ordered source-target pair $(X,Y)$ under parameters $\theta=\{w^{(b)},w^{(p)},a,c,k\}$, $B(X,Y)$ is a base similarity score learned from metric values, and $G(X,Y)$ is a gated prior modulation learned from source-wise metric ranks.
The parameter vectors $w^{(b)}$ and $w^{(p)}$ are the base and prior metric weights, while $a$, $c$, and $k$ are the gate modulation strength, threshold, and sharpness.
The following subsections define these two components and the parameters $\theta$ used to combine them.
Figure~\ref{fig:disc_architecture} provides an overview of DISC, where candidate metrics are learned as complementary base and prior evidence, and a gated aggregation converts them into the final dataset-similarity score.
We use an ordered-pair formulation because downstream transfer is directed: a model trained on $X$ and evaluated on $Y$ need not have the same error as a model trained on $Y$ and evaluated on $X$.
DISC learns from directed source-target pairs.

The desired association is monotone with error.
For a fixed source dataset $X$, if target dataset $Y$ produces a larger downstream error than target dataset $Z$, then a useful distance should tend to assign a larger score to $(X,Y)$ than to $(X,Z)$, i.e., $S_\theta(X,Y) > S_\theta(X,Z)$.
We do not require exact error prediction.
This pairwise intuition motivates learning a score whose magnitude is positively aligned with source-to-target transfer error.

\subsection{Base and Prior Metric Selection}
\label{sec:method_selection}
In this subsection, we describe how DISC converts raw metric distances into base and prior values, and how it learns which metrics should contribute to each role.
Different metrics have different scales.
For a candidate set $\mathcal{A}$, each source row and metric is standardized to obtain $z_m(X,Y)$, the normalized value of raw distance $s_m(X,Y)$ within the source row of $X$.
We also compute the source-wise percentile rank within the same candidate set:
\begin{equation}
r_m(X,Y)=
\operatorname{RankPct}_{Z\in\mathcal{A}\setminus\{X\}}
\left(s_m(X,Y)\right)
\in[0,1].
\end{equation}
Here, $\operatorname{RankPct}$ returns the percentile position of $s_m(X,Y)$ among the distances from source $X$ to candidate targets $Z$, with larger values indicating relatively larger distance.
The value feature $z_m$ preserves normalized metric magnitude.
The rank feature $r_m$ preserves relative source-wise position and is less sensitive to scale.
DISC uses these two feature types for different roles.
Standardized metric values provide the base similarity value, while source-wise ranks provide a prior value over relative source-target evidence.
This distinction is important in heterogeneous time-series repositories.
For example, distributional metrics may produce large absolute values for high-scale datasets, whereas rank signals still indicate which targets are relatively close to a given source.

The base mixture selects and weights metrics that define the primary score:
\begin{equation}
B(X,Y)=\sum_{m\in\mathcal{M}} w^{(b)}_m z_m(X,Y),
\quad
w^{(b)}=\mathrm{softmax}(\alpha/\tau).
\label{eq:base}
\end{equation}
Here, $w^{(b)}_m$ is the learned base weight for metric $m$, $\alpha$ is the base-logit vector before softmax normalization, and $\tau$ is the softmax temperature.
The learned logits $\alpha$ determine which metrics contribute to the base score.
The temperature $\tau$ controls the concentration of the softmax mixture.
The prior mixture independently selects and weights metrics that provide supplementary rank information:
\begin{equation}
P(X,Y)=\sum_{m\in\mathcal{M}} w^{(p)}_m r_m(X,Y),
\quad
w^{(p)}=\mathrm{softmax}(\beta/\tau).
\label{eq:prior}
\end{equation}
Here, $P(X,Y)$ is the rank-based prior signal, $w^{(p)}_m$ is the learned prior weight for metric $m$, and $\beta$ is the prior-logit vector.
The learned logits $\beta$ determine which metrics contribute to the prior signal.
Separating the base and prior mixtures allows DISC to use one set of metrics to define the main distance while using another to provide source-wise relative evidence.

\subsection{Gated Integration}
\label{sec:method_gate}
In this subsection, we describe how the prior metric modulates the base metric to instantiate the overall score in Eq.~\ref{eq:disc_overview}.
DISC combines the base and prior metrics with a controlled gate.
The base mixture provides the main standardized score, while the prior mixture provides relative evidence about which target datasets look distant from a source under rank-based metric views.
We therefore use the prior as a multiplicative interaction with the base score rather than as a standalone additive distance.
The multiplier starts at one and changes the magnitude of the base score according to the learned prior signal.
The prior signal modulates the base through
\begin{equation}
G(X,Y)=1+a\sigma\left(k(P(X,Y)-c)\right),
\label{eq:gate}
\end{equation}
where $\sigma(\cdot)$ is the sigmoid function.
The three gate parameters have direct semantic roles.
The modulation parameter $a$ controls the strength of the multiplicative interaction, the threshold $c$ determines the prior-rank level at which the interaction changes, and the sharpness $k$ controls whether this transition is gradual or close to a hard cutoff.
These parameters are needed because different downstream tasks may require different degrees of prior influence: some tasks benefit from broad rank-based modulation across many pairs, while others require the gate to respond mainly to high-prior, clearly distant pairs.
Learning these parameters lets DISC decide whether a downstream task benefits from weak, broad prior modulation or from a sharper rank-conditioned interaction.
The final DISC score uses this gate to adjust the base score, as shown in Eq.~\ref{eq:disc_overview}.
This form keeps the base score interpretable while allowing rank evidence from other metrics to interact with it.
Because $B(X,Y)$ is computed from standardized metric values, it is not constrained to be positive.
Accordingly, the gate is interpreted as a learned multiplicative modulation of the base score, not as a standalone monotone penalty.

We constrain $a$ with a sigmoid transform and a maximum modulation strength.
We constrain $c$ to $[0,1]$.
The gate sharpness $k$ is learned within a bounded interval.
In the final run, the bound is $[1,55]$.
This range is centered around the original fixed sharpness value of 40 but prevents the gate from becoming arbitrarily hard.

\subsection{Training Objective and Procedure}
\label{sec:method_training}
In this subsection, we present the loss function and training procedure used to learn the metric weights and gate parameters for each downstream task.
Let $E(X,Y)$ denote the downstream error for a source-target pair.
These downstream errors are used as calibration signals for learning the aggregation parameters $\theta$, including the base weights $w^{(b)}$, prior weights $w^{(p)}$, and gate parameters $a$, $c$, and $k$.
Once $\theta$ is fixed, the DISC score $S_\theta(X,Y)$ is computed from similarity features alone and does not require the downstream error of that pair.
For an evaluation set $\mathcal{A}$ and a set of source datasets $\mathcal{S}\subseteq\mathcal{A}$, where $\mathcal{A}$ defines the candidate datasets compared in the current stage and $\mathcal{S}$ defines the source rows being averaged, the source-wise Pearson correlation is
\begin{equation}
\rho_{\mathrm{src}}=
\frac{1}{|\mathcal{S}|}
\sum_{X\in\mathcal{S}}
\rho\left(
\{S_\theta(X,Y)\}_{Y\in\mathcal{A}\setminus\{X\}},
\{E(X,Y)\}_{Y\in\mathcal{A}\setminus\{X\}}
\right).
\label{eq:source_corr}
\end{equation}
Here, $\rho(\cdot,\cdot)$ denotes Pearson correlation, and each term compares the DISC scores and downstream errors within one fixed source row.
We also compute global Pearson correlation $\rho_{\mathrm{global}}$ over all source-target pairs.
The training objective is
\begin{equation}
\mathcal{J}(\theta)=
(1-\lambda)\rho_{\mathrm{src}}+\lambda\rho_{\mathrm{global}},
\label{eq:objective}
\end{equation}
where $\lambda=0.10$ controls the trade-off between source-wise and global alignment, giving primary weight to source-wise correlation while retaining a small global-pair alignment term.
Here, $\rho_{\mathrm{src}}$ measures the average within-source alignment, and $\rho_{\mathrm{global}}$ measures alignment over all directed source-target pairs.
The optimized loss is
\begin{equation}
\mathcal{L}=
-\mathcal{J}(\theta)
+\eta H(w^{(b)},w^{(p)})
+\mu\langle w^{(b)},w^{(p)}\rangle.
\label{eq:loss}
\end{equation}
The entropy term $H(w^{(b)},w^{(p)})$ regularizes the two softmax weight vectors and encourages sparse metric usage.
The inner product $\langle w^{(b)},w^{(p)}\rangle$ measures overlap between the base and prior metric mixtures, so the overlap term discourages the two mixtures from selecting the same metrics.
The coefficients $\eta=0.006$ and $\mu=0.02$ control the strengths of the entropy and overlap regularizers, respectively.
All parameters are optimized with Adam.

Training proceeds independently for each downstream task.
First, we compute candidate metric matrices from dataset samples.
Second, we build directed source-target entries for the current optimization or evaluation stage.
Third, we attach standardized metric values, source-wise ranks, and the corresponding downstream error entries for that stage.
Fourth, we optimize the DISC parameters on the training datasets and select the checkpoint with the best validation source-wise correlation.
Finally, we freeze the learned parameters and report test-split source-wise correlation.
Thus, downstream transfer results are used to calibrate how metrics should be combined, while the reported DISC scores on the test split are produced from the learned aggregation of similarity metrics.

The optimization uses random restarts.
Each restart initializes the base and prior logits near zero, which starts from a nearly uniform mixture.
The scalar gate parameters are initialized so that the gate is smooth at the beginning of training.
We keep the validation criterion identical to the reporting metric, except that validation datasets replace test datasets.
This keeps the validation criterion consistent with the final reporting metric.
The learning rate controls the Adam step size, epochs set the maximum number of training iterations, patience stops a restart when validation alignment no longer improves, and restarts repeat the optimization from different random initializations.
The softmax temperature $\tau$ controls how concentrated the metric weights become.
The maximum gate modulation bounds $a$, the gate threshold range bounds $c$, and the gate sharpness range bounds $k$.
Table~\ref{tab:hyperparameters} lists the optimization and regularization settings used for all reported DISC runs.

\begin{table}[t]
\centering
\caption{DISC training hyperparameters used in the main experiment.}
\label{tab:hyperparameters}
\begin{tabular}{lc|lc}
\toprule
\textbf{Hyperparameter} & \textbf{Value} & \textbf{Hyperparameter} & \textbf{Value} \\
\midrule
Optimizer & Adam & Learning rate & 0.03 \\
Epochs & 500 & Patience & 100 \\
Restarts & 8 & Softmax temperature $\tau$ & 0.55 \\
Max gate modulation & 3.0 & Gate threshold range & $[0,1]$ \\
Gate sharpness range & $[1,55]$ & Global-correlation weight $\lambda$ & 0.10 \\
Entropy weight $\eta$ & 0.006 & Overlap weight $\mu$ & 0.02 \\
\bottomrule
\end{tabular}
\end{table}

\section{Experiments}
\label{sec:experiments}
We structure the experiments around 3 research questions.

\textbf{RQ1.} Do single similarity metrics consistently predict downstream errors?

\textbf{RQ2.} Does DISC show higher alignment than the best single metric on the reported split?

\textbf{RQ3.} Does learned aggregation outperform non-learned parameter search?

\begin{table*}[t]
\centering
\caption{Correlation between similarity value and downstream task error for single metrics and DISC. OT is the reference; parentheses show gains over OT, bold marks the best value, and italics mark the second-best single metric.}
\label{tab:main_result_part1}
\setlength{\tabcolsep}{4pt}
\begin{tabular}{lcccccc}
\toprule
\multirow{2}{*}[-0.8ex]{\textbf{Task}} & \multicolumn{6}{c}{\textbf{Correlation (gain over OT)}} \\
\cmidrule(lr){2-7}
& \textbf{OT} & \textbf{WSD} & \textbf{MMD} & \textbf{MAD} & \textbf{DBA-DTW} & \textbf{DISC (ours)} \\
\midrule
OOD-TM & -0.007 (ref.) & 0.445 ($+0.453\uparrow$) & -0.057 ($-0.050\downarrow$) & \second{0.580 ($+0.588\uparrow$)} & -0.016 ($-0.008\downarrow$) & \best{0.601 ($+0.609\uparrow$)} \\
OOD-LL & 0.125 (ref.) & 0.361 ($+0.235\uparrow$) & 0.295 ($+0.169\uparrow$) & 0.231 ($+0.105\uparrow$) & \second{0.618 ($+0.492\uparrow$)} & \best{0.697 ($+0.571\uparrow$)} \\
TL-TM & 0.213 (ref.) & 0.516 ($+0.303\uparrow$) & 0.303 ($+0.091\uparrow$) & 0.505 ($+0.292\uparrow$) & 0.134 ($-0.079\downarrow$) & \best{0.617 ($+0.404\uparrow$)} \\
TL-LL & 0.080 (ref.) & 0.011 ($-0.069\downarrow$) & \second{0.435 ($+0.355\uparrow$)} & 0.146 ($+0.066\uparrow$) & 0.302 ($+0.222\uparrow$) & \best{0.479 ($+0.399\uparrow$)} \\
Avg & 0.103 (ref.) & 0.333 ($+0.231\uparrow$) & 0.244 ($+0.141\uparrow$) & 0.366 ($+0.263\uparrow$) & 0.259 ($+0.157\uparrow$) & \best{0.598 ($+0.496\uparrow$)} \\
\bottomrule
\end{tabular}
\end{table*}

\begin{table*}[t]
\centering
\caption{Correlation between similarity value and downstream task error for single metrics and DISC (cont.). OT is the reference; parentheses show gains over OT, bold marks the best value, and italics mark the second-best single metric.}
\label{tab:main_result_part2}
\scriptsize
\setlength{\tabcolsep}{3pt}
\begin{tabular}{lccccccc}
\toprule
\multirow{2}{*}[-0.8ex]{\textbf{Task}} & \multicolumn{7}{c}{\textbf{Correlation (gain over OT)}} \\
\cmidrule(lr){2-8}
& \textbf{OT} & \textbf{DBA-ED} & \textbf{DBA-LCSS} & \textbf{PCA-DTW} & \textbf{PCA-ED} & \textbf{PCA-LCSS} & \textbf{DISC (ours)} \\
\midrule
OOD-TM & -0.007 (ref.) & -0.016 ($-0.008\downarrow$) & 0.000 ($+0.007\uparrow$) & 0.381 ($+0.389\uparrow$) & 0.416 ($+0.423\uparrow$) & 0.048 ($+0.055\uparrow$) & \best{0.601 ($+0.609\uparrow$)} \\
OOD-LL & 0.125 (ref.) & \second{0.618 ($+0.492\uparrow$)} & 0.052 ($-0.074\downarrow$) & 0.577 ($+0.452\uparrow$) & 0.580 ($+0.454\uparrow$) & 0.461 ($+0.336\uparrow$) & \best{0.697 ($+0.571\uparrow$)} \\
TL-TM & 0.213 (ref.) & 0.134 ($-0.079\downarrow$) & 0.026 ($-0.187\downarrow$) & 0.482 ($+0.269\uparrow$) & \second{0.537 ($+0.325\uparrow$)} & -0.025 ($-0.238\downarrow$) & \best{0.617 ($+0.404\uparrow$)} \\
TL-LL & 0.080 (ref.) & 0.302 ($+0.222\uparrow$) & 0.000 ($-0.080\downarrow$) & 0.113 ($+0.033\uparrow$) & 0.078 ($-0.002\downarrow$) & 0.107 ($+0.027\uparrow$) & \best{0.479 ($+0.399\uparrow$)} \\
Avg & 0.103 (ref.) & 0.259 ($+0.157\uparrow$) & 0.019 ($-0.083\downarrow$) & 0.388 ($+0.286\uparrow$) & \second{0.403 ($+0.300\uparrow$)} & 0.148 ($+0.045\uparrow$) & \best{0.598 ($+0.496\uparrow$)} \\
\bottomrule
\end{tabular}
\end{table*}

\subsection{Datasets and Splits}
We use 46 datasets from GluonTS~\cite{alexandrov2020gluonts}.
For similarity computation, each dataset is represented by 100 sampled windows of length 16.
The final evaluation uses a shared split across the four forecasting tasks: 24 training datasets, 8 validation datasets, and 14 test datasets.
The training datasets provide the calibration observations used to optimize DISC parameters.
Validation datasets are used to select the calibrated aggregation setting.
All reported numbers are computed on the designated test split.
The 14 test datasets define 14 source rows, so evaluation repeatedly asks whether DISC explains transfer behavior for each newly source dataset against the other targets.
The shared split controls for dataset-selection effects across tasks.
Each downstream task uses the same dataset split, so differences in results reflect task behavior rather than changes in the evaluated dataset set.

\subsection{Downstream Tasks}
We evaluate forecasting performance using two foundation models, \textbf{Time-MoE}~\cite{shi2025time} and \textbf{Lag-Llama}~\cite{rasul2023lag}, under out-of-distribution and transfer-learning protocols.
In the \textbf{out-of-distribution (OOD)} protocol, a model is fine-tuned on a source dataset and evaluated on target datasets without target adaptation.
In the \textbf{transfer-learning (TL)} protocol, the model is first fine-tuned on a source dataset and then adapted with target-side data before final target evaluation.
With 46 datasets, the OOD protocol fine-tunes 46 source models for each foundation model and evaluates them on $46\times45=2{,}070$ directed source-target pairs per foundation model after excluding self-pairs.
The TL protocol starts from the same source fine-tuned models, but additionally adapts each source model to each target dataset, resulting in 2,070 target-adapted source-target runs per foundation model.
This gives four downstream tasks: Out-of-distribution Time-MoE, Out-of-distribution Lag-Llama, Transfer learning with Time-MoE, and Transfer learning with Lag-Llama.
The downstream error matrix records the loss for each source-target pair.

\subsection{Similarity Metrics}
We compare multiple similarity configurations that encode different assumptions about time-series dataset relationships:
\begin{itemize}
    \item \textit{Dataset-level metrics}: \textit{Wasserstein Distance (WSD)}~\cite{chen2025measuring} captures distributional mismatch through moment-based geometry, \textit{Maximum Mean Discrepancy (MMD)}~\cite{gretton2012kernel} compares empirical distributions through kernel mean embeddings, \textit{Optimal Transport (OT)}~\cite{cuturi2013sinkhorn,gabriel2019computational} measures the minimum transport cost between empirical samples, and \textit{Match-and-Deform (MAD)}~\cite{painblanc2023match} combines sample matching with temporal alignment.
    \item \textit{Dataset reducers}: \textit{Dynamic Time Warping Barycenter Averaging (DBA)}~\cite{petitjean2011global} summarizes each dataset as a DTW barycenter, while \textit{Principal Component Analysis (PCA)}~\cite{pearson1901liii} summarizes each dataset through a dominant projection. These reducers are not standalone similarity metrics in our comparison; they provide representative sequences for series-level distances.
    \item \textit{Series-level distances}: \textit{Dynamic Time Warping (DTW)}~\cite{sakoe1978dynamic}, \textit{Euclidean Distance}~\cite{faloutsos1994fast,ding2008querying}, and \textit{Longest Common Subsequence (LCSS)}~\cite{hirschberg1977algorithms,vlachos2002discovering} compare the DBA or PCA representatives through warping, aligned shape distance, or subsequence matching.
\end{itemize}
Combining the DBA and PCA reducers with DTW, Euclidean distance, and LCSS yields six reducer-based configurations: DBA-DTW, DBA-ED, DBA-LCSS, PCA-DTW, PCA-ED, and PCA-LCSS.
Together with the four dataset-level metrics, these configurations cover complementary views of source-target relationships, allowing the experiments to test whether their usefulness varies across models, tasks, and adaptation settings.

\begin{figure*}[t]
\centering
\includegraphics[width=0.63\linewidth]{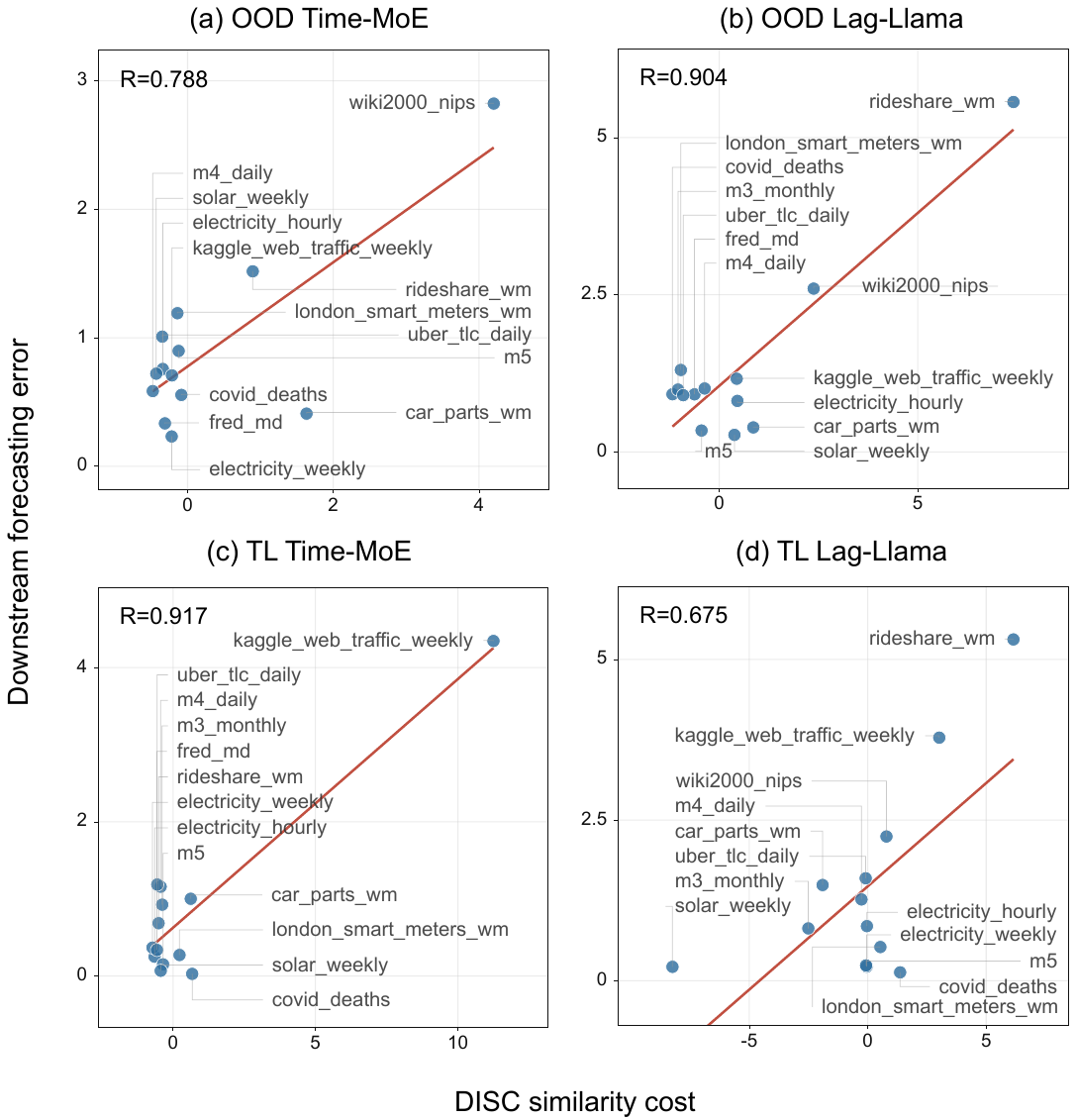}
\caption{Source-wise DISC examples for RQ2. Each panel fixes the test source dataset with the highest DISC source-wise Pearson correlation for one downstream task: (a) \texttt{m3\_monthly}, (b) \texttt{electricity\_weekly}, (c) \texttt{wiki2000\_nips}, and (d) \texttt{fred\_md}. The x-axis is the learned DISC similarity score and the y-axis is downstream forecasting error. Upward trends show that datasets assigned larger DISC scores tend to produce larger errors.
\label{fig:rq2_best_source_examples}}
\end{figure*}

\subsection{Evaluation Protocol}
For each source dataset in the test split, we compute Pearson correlation between its similarity distances to the other test datasets and the corresponding downstream errors.
The purpose is to test whether a pre-adaptation DISC score is comparable to the forecasting error observed after using source dataset $X$ on target dataset $Y$.
This evaluates source-conditioned alignment between dataset similarity and downstream behavior, rather than target-wise top-$k$ source retrieval.
With 14 test datasets, each source-level correlation uses 13 target points.
The reported score is the mean of the 14 source-level correlations, treating each source row as a separate evaluation case.
Higher positive values indicate stronger alignment between dataset distance and downstream error.
Because the downstream error $E(X,Y)$ also reflects the intrinsic forecasting difficulty of target dataset $Y$, the reported correlations measure alignment with observed source-target forecasting behavior rather than isolating transfer compatibility alone.
In this analysis, target difficulty is part of the realized downstream cost that a source-target screening score should anticipate, not a quantity that DISC is claimed to decompose from pair-specific compatibility.
This source-wise definition prevents a few high-variance sources from dominating the global result.
It evaluates whether a fixed source dataset induces a meaningful score-error association over target datasets.
Because transfer errors are directed, we use this metric specifically to evaluate source-row transfer alignment.
The source-wise view asks whether each source row provides a meaningful local explanation of downstream transfer behavior.
This evaluation is therefore a correlation-alignment analysis, not a direct source-selection benchmark.

\begin{table*}[t]
\centering
\caption{DISC compared with the best single metric for each downstream task on the reported split. R denotes Pearson correlation between similarity score and downstream task error; higher values indicate stronger alignment.}
\label{tab:best_single}
\begin{tabular}{cccccc}
\toprule
\multicolumn{2}{c}{\textbf{Task}} & \multirow{2}{*}[-0.7ex]{\textbf{Best single}} & \multirow{2}{*}[-0.7ex]{\textbf{Single R $\uparrow$}} & \multirow{2}{*}[-0.7ex]{\textbf{DISC R $\uparrow$}} & \multirow{2}{*}[-0.7ex]{\textbf{Relative change $\uparrow$}} \\
\cmidrule(lr){1-2}
\textbf{Protocol} & \textbf{Model} & & & & \\
\midrule
\multirow{2}{*}{Out-of-distribution} & Time-MoE & MAD & 0.580 & 0.601 & +3.55\% \\
& Lag-Llama & DBA-DTW & 0.618 & 0.697 & +12.84\% \\
\multirow{2}{*}{Transfer learning} & Time-MoE & PCA-ED & 0.537 & 0.617 & +14.75\% \\
& Lag-Llama & MMD & 0.435 & 0.479 & +10.15\% \\
\midrule
\multicolumn{2}{c}{Average} & -- & 0.543 & 0.598 & +10.29\% \\
\bottomrule
\end{tabular}
\end{table*}

\begin{table*}[t]
\centering
\caption{Learned DISC aggregation for each downstream task. Base and prior columns list selected metrics with learned softmax weights in parentheses; metrics below 0.01 are omitted. Prior modulation reports the mean and range of the positive gate factor multiplied with the base score on test-split pairs.}
\label{tab:learned_components}
\begingroup
\scriptsize
\setlength{\tabcolsep}{3pt}
\newcolumntype{C}[1]{>{\centering\arraybackslash}p{#1}}
\begin{tabular}{C{0.13\linewidth}C{0.10\linewidth}p{0.22\linewidth}p{0.34\linewidth}C{0.12\linewidth}}
\toprule
\multicolumn{2}{c}{\textbf{Task}} & \multirow{2}{*}[-0.8ex]{\textbf{Base metrics}} & \multirow{2}{*}[-0.7ex]{\textbf{Prior metrics}} & \multirow{2}{*}[-0.7ex]{\textbf{Prior modulation}} \\
\cmidrule(lr){1-2}
\textbf{Protocol} & \textbf{Model} & & & \\
\midrule
\multirow{2}{*}[-1.8ex]{Out-of-distribution} & Time-MoE & \begin{tabular}[c]{@{}l@{}}DBA-LCSS (.549), MAD (.402)\\ Wasserstein (.022)\end{tabular} & \begin{tabular}[c]{@{}l@{}}PCA-ED (.717), Wasserstein (.173)\\ DBA-ED (.045), DBA-DTW (.037), PCA-LCSS (.012)\end{tabular} & 2.26$\times$ [1.00, 3.35] \\
\cmidrule(lr){2-5}
& Lag-Llama & \begin{tabular}[c]{@{}l@{}}PCA-DTW (.612)\\ PCA-LCSS (.383)\end{tabular} & \begin{tabular}[c]{@{}l@{}}PCA-LCSS (.935)\\ DBA-LCSS (.062)\end{tabular} & 1.21$\times$ [1.00, 3.79] \\
\midrule
\multirow{2}{*}[-1.8ex]{Transfer learning} & Time-MoE & \begin{tabular}[c]{@{}l@{}}MAD (.916)\\ Wasserstein (.067)\end{tabular} & \begin{tabular}[c]{@{}l@{}}PCA-ED (.719), DBA-ED (.180)\\ DBA-DTW (.076), MMD (.017)\end{tabular} & 1.17$\times$ [1.00, 3.81] \\
\cmidrule(lr){2-5}
& Lag-Llama & \begin{tabular}[c]{@{}l@{}}MMD (.833)\\ DBA-ED (.159)\end{tabular} & \begin{tabular}[c]{@{}l@{}}DBA-DTW (.362), MAD (.324)\\ PCA-ED (.312)\end{tabular} & 2.89$\times$ [1.00, 3.95] \\
\bottomrule
\end{tabular}
\endgroup
\end{table*}

\subsection{RQ1: Single Similarity Configurations Are Task-Dependent}
Tables~\ref{tab:main_result_part1} and~\ref{tab:main_result_part2} report all single similarity configurations and DISC, Optimal Transport is used as the reference metric, and the parenthesized values show how much each configuration increases or decreases correlation relative to Optimal Transport.
Single configurations are inconsistent across tasks.
MAD is the strongest single metric for OOD-Time-MoE.
DBA-DTW and DBA-ED are strongest for OOD-Lag-Llama.
PCA-ED is strongest for TL-Time-MoE.
MMD is strongest for TL-Lag-Llama.
The best single metric changes with the downstream model and protocol.

The average single-metric result also shows the limitation of manual selection.
PCA-ED has the highest average among single metrics, with 0.403.
The per-task oracle, which selects the best single metric separately for each task, reaches 0.543.
DISC exceeds both, reaching 0.598.
The gap between the best average single metric and the per-task oracle is large.
This means that a metric selected once for all tasks is not sufficient.
The remaining gap between the per-task oracle and DISC shows that aggregation can improve even when the strongest single metric is selected separately for each task.

\subsection{RQ2: DISC Shows Gains Through Task-Specific Aggregation}
Table~\ref{tab:best_single} compares DISC with the best single metric for each task.
On the reported split, DISC is higher than the best single metric in all four task columns.
The average relative change is 10.29\%.
The largest margins occur for OOD-Lag-Llama and TL-Time-MoE, where the best single metric leaves substantial unexplained structure.
The OOD-Time-MoE margin is modest because MAD is already strong on this task.
We therefore interpret this case as a small directional gain rather than as evidence of a large effect.
The TL-Lag-Llama task is the weakest absolute task for both single metrics and DISC, but DISC is still higher than MMD by 10.15\%.
These results are based on one shared train-validation-test split, but the test score is averaged over 14 source-row evaluations and repeated across four downstream settings.
We therefore interpret the evidence as source-row transfer alignment under the shared split, rather than as a repeated-split uncertainty analysis over all possible dataset partitions.

Figure~\ref{fig:rq2_best_source_examples} provides source-level examples of how DISC separates target datasets across different downstream tasks.
We show one representative fixed source from each forecasting setting, chosen as the source with the strongest DISC source-wise correlation, and compare its learned DISC scores with the downstream errors of the remaining test-split target datasets.
Across the four panels, targets with larger DISC scores generally incur larger forecasting errors.
This pattern indicates that DISC does not only improve the aggregate correlation in Table~\ref{tab:best_single}; it also produces visible local score-error alignment within individual source rows.
Such source-wise separation is useful for analyzing how transfer error changes from a fixed source to different target datasets.
Thus, the figure supports the intended source-conditioned interpretation: DISC scores are compared with the realized downstream errors of using the fixed source on each target.
The panels are illustrative examples, while the reported RQ2 scores average this source-wise alignment over all 14 source rows.

The observed gain comes from task-specific metric selection and integration rather than from a fixed global metric choice.
Table~\ref{tab:learned_components} reports the sparse components of the learned base and prior mixtures, their learned weights, and the resulting prior-gate modulation factor on test-split pairs.
The base weights indicate how DISC combines metric values to form the primary similarity score.
The prior weights indicate which metric ranks are used to form the gate signal.
The prior modulation factor summarizes how strongly the prior signal interacts with the learned base score on the test pairs.
OOD-Time-MoE uses MAD with DBA-LCSS and Wasserstein-related information.
OOD-Lag-Llama relies on PCA-based shape summaries.
TL-Time-MoE uses MAD and Wasserstein in the base and uses PCA-ED, DBA-ED, DBA-DTW, and MMD as priors.
TL-Lag-Llama uses MMD and DBA-ED as the base and uses DBA-DTW, MAD, and PCA-ED as priors.
The learned components match the task-dependent behavior in Tables~\ref{tab:main_result_part1} and~\ref{tab:main_result_part2}.
They also show why the method should not be described as improving one particular metric.
For example, DISC uses MMD as part of the TL-Lag-Llama base, but not as the OOD-Lag-Llama base.
It uses PCA-LCSS as both a base and prior signal in OOD-Lag-Llama, but uses PCA-ED as a prior in the Time-MoE tasks.
The learned structure is therefore task-specific rather than metric-specific.

\subsection{RQ3: Learned Aggregation Versus Manual Search}
We compare DISC with a non-learned manual search over the same base-prior aggregation family.
This comparison tests whether the best similarity-metric combination found from previously observed datasets can be directly reused for unseen datasets.
The manual search samples sparse base mixtures, sparse prior mixtures, and gate parameters, then selects the best configuration using the training and validation datasets.
The selected configuration is applied once to the same test split used in Tables~\ref{tab:main_result_part1} and~\ref{tab:main_result_part2}.
This keeps the manual-search baseline under the same selection protocol as DISC.
It therefore asks whether broad manual parameter search can replace learned aggregation under the same evaluation setting.

The answer is no.
Table~\ref{tab:ablation} reports the task-wise test-split comparison.
The extended manual search evaluates approximately 7.9 million candidate configurations per task, but its average test-split correlation is 0.291.
DISC reaches 0.598 under the same test-split protocol.
The manual search also remains below DISC on every task, with test correlations of -0.066, 0.571, 0.316, and 0.344 for the four settings.
The learned combinations remain effective on the fourteen test datasets, indicating that the selected metric weights and gate behavior still align with newly evaluated source-target relationships.
The N/A gain in Table~\ref{tab:ablation} occurs for OOD-Time-MoE because the manual-search baseline is negative, making a relative percentage gain misleading.
Its learnable parameters provide a reliable way to select metric weights and gate behavior that generalize from the training and validation setting to the test split.

\begin{table}[t]
\centering
\caption{Task-wise ablation against non-learned manual search. R denotes Pearson correlation between similarity distance and downstream task error. Manual R and DISC R are test-split correlations after parameter selection on training and validation datasets; Gain is DISC's relative improvement, with N/A used for a negative manual baseline.}
\label{tab:ablation}
\begin{tabular}{lccc}
\toprule
\textbf{Task} & \textbf{Manual R} & \textbf{DISC R} & \textbf{Gain} \\
\midrule
OOD-TM & -0.066 & 0.601 & N/A \\
OOD-LL & 0.571 & 0.697 & +21.98\% \\
TL-TM & 0.316 & 0.617 & +95.13\% \\
TL-LL & 0.344 & 0.479 & +39.32\% \\
\midrule
Avg & 0.291 & 0.598 & +105.38\% \\
\bottomrule
\end{tabular}
\end{table}

\subsection{Discussion: Dataset Characteristics and Metric Reliability}
We further examine the dataset characteristics associated with weaker single-metric explanations of downstream behavior.
The pattern is visible in the same $N=100$, $L=16$ sampled windows used for similarity computation, as summarized in Table~\ref{tab:diagnostics}.
Here, $N$ is the number of sampled windows per dataset and $L$ is the sampled-window length.
These patterns suggest that the issue is not only metric design; overly smooth, flat, or low-diversity sampled representations can also limit how well similarity metrics explain source-target relationships.

The results show that similarity metrics are not interchangeable, rank information can complement standardized metric values, and sampling choices affect similarity estimation when short windows miss seasonality or overrepresent flat regimes.
They also clarify that Table~\ref{tab:best_single} is a favorable baseline because it selects the best single metric per task from the test result table, a choice unavailable before evaluation.
DISC improving over this reference suggests that learned aggregation does more than choose the obvious best metric.

\begin{table}[t]
\centering
\caption{Sample-window diagnostics for low- and high-correlation groups. Lag-1 is adjacent-step autocorrelation, zero is the zero-valued sample fraction, and diversity is mean pairwise distance between z-normalized windows.}
\label{tab:diagnostics}
\begin{tabular}{lccc}
\toprule
\textbf{Group} & \textbf{Lag-1} & \textbf{Zero} & \textbf{Diversity} \\
\midrule
Low corr. datasets & 0.841 & 0.191 & 3.935 \\
High corr. datasets & 0.556 & 0.061 & 5.115 \\
\bottomrule
\end{tabular}
\end{table}

\section{Conclusion}
We introduced DISC, a learnable framework for time-series dataset similarity.
DISC integrates dataset-level and reducer-based similarity configurations through a calibrated base mixture and a rank-conditioned prior modulation, producing a combined similarity score that better separates source-target relationships across downstream tasks.
On the reported split across four forecasting tasks with Time-MoE and Lag-Llama, DISC is higher than the best single metric in every task column and increases the average correlation by 10.29\%.
The study shows that source-target transfer alignment is task-dependent and cannot be reliably solved by one fixed metric.
It also shows that short, flat, or low-diversity sampled windows can weaken single-metric correlation.
These results support learned metric aggregation as a practical way to turn calibration transfer observations and heterogeneous similarity signals into a task-aware score for time-series source-target transfer assessment.

\bibliographystyle{IEEEtran}
\bibliography{references}

\end{document}

%% file: preamble.tex
\usepackage{wasysym}

\usepackage{algorithm}
\usepackage{algorithmic}
\usepackage{tcolorbox}
\usepackage{fontawesome5}
\usepackage{pifont}
\usepackage{tikz}
\usepackage{tcolorbox}
\tcbuselibrary{skins}
\newtcolorbox{PromptTextBox}[1]{%
  colback=gray!5,
  colframe=black,
  boxrule=0.5pt,
  arc=2mm,
  width=\linewidth,
  title={#1},
  fonttitle=\bfseries,
  enhanced,
  left=2mm,
  right=2mm,
  top=1mm,
  bottom=1mm,
  boxsep=1mm
}

\usepackage{enumitem}
\setlist[itemize]{leftmargin=*} % removes indent
\usepackage{CJKutf8}

\usepackage{tabularx}
\usepackage{xcolor}
\usepackage{colortbl}

\definecolor{googleblue}{RGB}{66,133,244}
\definecolor{googlered}{RGB}{219,68,55}
\definecolor{googlegreen}{RGB}{15,157,88}
\definecolor{googlepurple}{RGB}{138,43,226}
\definecolor{lightred}{RGB}{255, 220, 219}
\definecolor{lightblue}{RGB}{204, 243, 255}
\definecolor{lightgreen}{RGB}{200, 247, 200}
\definecolor{lightpurple}{RGB}{230,230,250}
\definecolor{lightyellow}{RGB}{242, 232, 99}
\definecolor{lighterblue}{RGB}{197, 220, 255}
\definecolor{lighterred}{RGB}{253, 249, 205}
\definecolor{lightyellow}{RGB}{207, 161, 13}
\definecolor{darkpurple}{RGB}{218, 210, 250}
\definecolor{darkred}{RGB}{255,198,196}
\definecolor{darkblue}{RGB}{172, 233, 252}

\usepackage{amsmath}
\usepackage{comment}
\usepackage{graphicx}
\usepackage{multirow}
\usepackage{booktabs}
\usepackage{caption}
\usepackage{array}
\usepackage{hyperref}

\usepackage{pifont} % for crossmark \ding{55}

\newcolumntype{H}{>{\setbox0=\hbox\bgroup}c<{\egroup}@{}}

\definecolor{thedarkblue}{RGB}{0,0,120} %104} % 180
\definecolor{mygreen}{RGB}{0,100,0} %104} % 180
\definecolor{mydarkblue}{rgb}{0,0.08,0.45} %ICML dark blue
\definecolor{darkblue}{rgb}{0,0.08,180}
\definecolor{orange}{HTML}{E6550D}
\colorlet{TufteRed}{red!80!black}

\definecolor{theblue}{RGB}{0,0,180}
\colorlet{thered}{TufteRed}

\providecommand{\eat}[1]{\ignorespaces}
\providecommand{\addressed}[1]{\ignorespaces}
\providecommand{\todoeat}[1]{\ignorespaces}
\providecommand{\supp}[1]{\ignorespaces} % content wrapped in \supp{} is for supp material

\DeclareMathOperator{\hugeE}{\mbox{\huge\raise-0.3ex\hbox{E}}}
\DeclareMathOperator{\p}{\mathbb{P}}
\DeclareMathOperator{\hugep}{\mbox{\huge\raise-0.3ex\hbox{$\p$}}}